\documentclass{article} 
\usepackage{iclr2021_conference,times}

\usepackage{amsmath,amsfonts,bm}

\def\eqref#1{equation~\ref{#1}}

\def\1{\bm{1}}

\DeclareMathAlphabet{\mathsfit}{\encodingdefault}{\sfdefault}{m}{sl}
\SetMathAlphabet{\mathsfit}{bold}{\encodingdefault}{\sfdefault}{bx}{n}

\usepackage{hyperref}
\usepackage{url}

\usepackage{caption}
\usepackage{graphicx}
\usepackage{multirow}
\usepackage{multicol}
\usepackage{amsmath}
\usepackage[para]{threeparttable}
\usepackage{booktabs}
\usepackage{tabularx}
\usepackage{makecell}

\usepackage{pifont}
\usepackage{mathabx}

\usepackage{subcaption}

\newcommand{\Ours}{\textsc{Reader-Distilled Retriever}}
\newcommand{\ours}{\textsc{RDR}}

\title{Is Retriever \\
Merely An Approximator of Reader?}

\author{Sohee Yang \& Minjoon Seo\\
NAVER Corp.\\
\texttt{\{sh.yang,minjoon.seo\}@navercorp.com} \\
}

\iclrfinalcopy 

\begin{document}

\maketitle

\begin{abstract}
The state of the art in open-domain question answering (QA) relies on an efficient retriever that drastically reduces the search space for the expensive reader. A rather overlooked question in the community is the relationship between the retriever and the reader, and in particular, if the whole purpose of the retriever is just a fast approximation for the reader. Our empirical evidence indicates that the answer is \emph{no}, and that the reader and the retriever are complementary to each other even in terms of accuracy only. We make a careful conjecture that the architectural constraint of the retriever, which has been originally intended for enabling approximate search, seems to also make the model more robust in large-scale search.  We then propose to distill the reader into the retriever so that the retriever absorbs the strength of the reader while keeping its own benefit. Experimental results show that our method can enhance the document recall rate as well as the end-to-end QA accuracy of off-the-shelf retrievers in open-domain QA tasks.

\end{abstract}

\section{Introduction}\label{sec:intro}
\begin{figure}[t] 
\centering
\begin{subfigure}[b]{0.32\textwidth}
\caption{One-Tower Model}
\includegraphics[width=\textwidth]{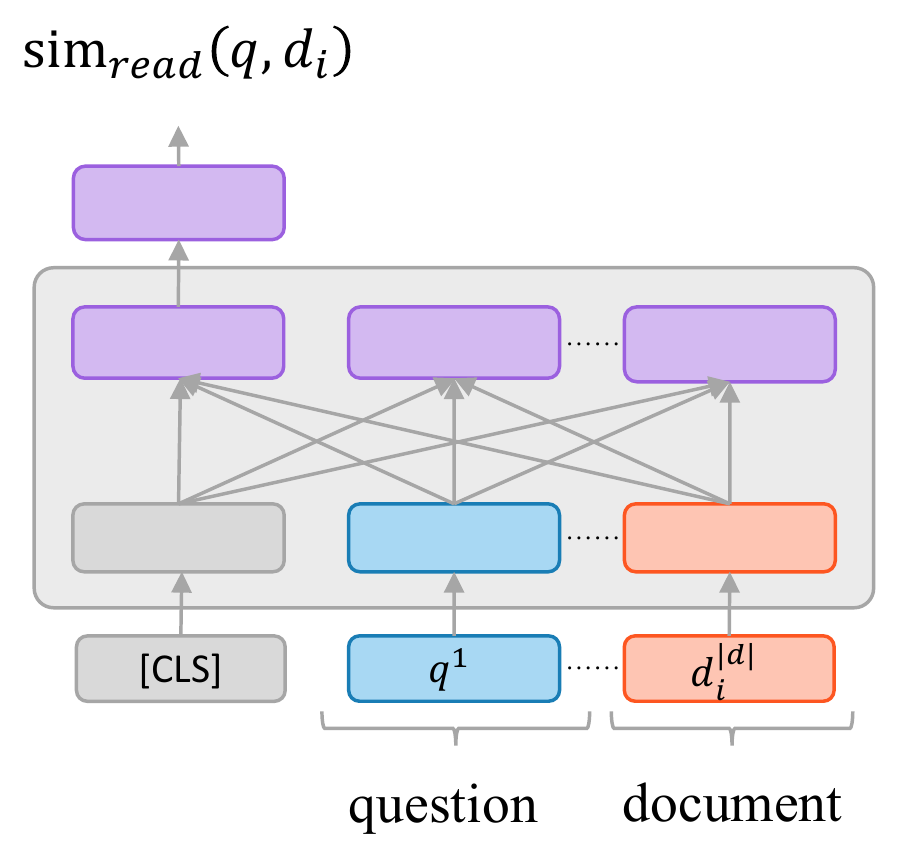}
\end{subfigure}
\begin{subfigure}[b]{0.62 \textwidth}
\caption{Two-Tower Model}
\includegraphics[width=\textwidth]{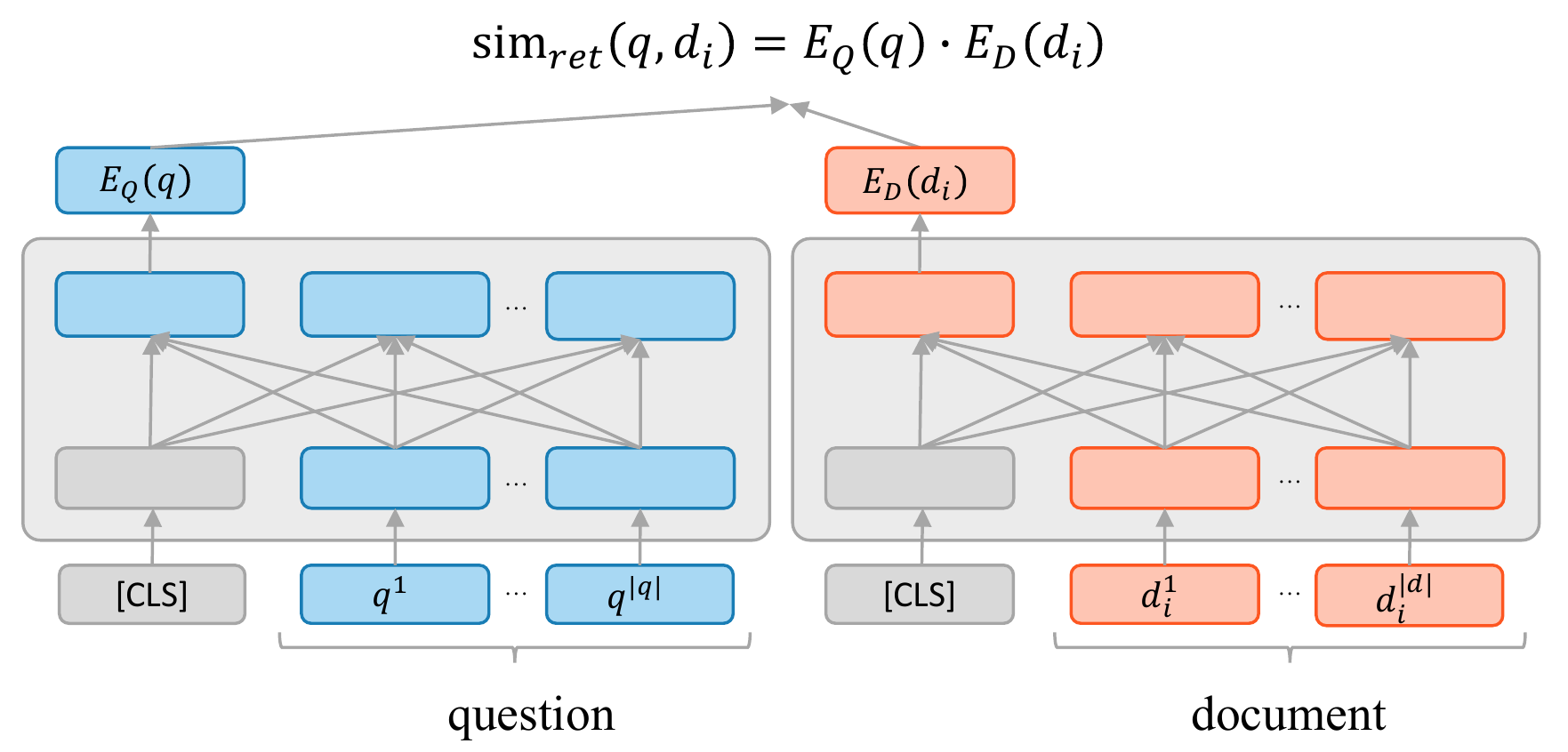}
\end{subfigure}
\caption{\small Comparison between the architectures of a one-tower model, e.g., reader, and a two-tower model, e.g., retriever, with BERT~\citep{bert}. The one-tower model takes the question and document together and jointly processes them throughout all the layers, while the two-tower architecture separately models the question and document whose outputs interact only at the inner product space of the final embeddings.}
\label{fig:teaser}
\end{figure}

The task of open-domain question answering can be defined as creating a model that takes a question and the knowledge source as the input and outputs the answer to the question. 
In this paper, we primarily focus on unstructured (text) knowledge data such as Wikipedia, and we do not consider structured sources such as Knowledge Graph.
In most cases~\citep{karpukhin2020dense, lewis2020retrieval, izacard2020leveraging}, since the (unstructured) knowledge data is so big, one first \emph{retrieves} a few relevant documents to the question from the knowledge data and then \emph{reads} the retrieved documents to obtain the answer.
For the retriever to quickly search over a large number of documents, its architecture is often constrained to be a \emph{two-tower} (Figure~\ref{fig:teaser}b), where the question and the documents are independently mapped to a common vector space. This way, fast sublinear-time approximation methods such as approximate nearest neighbor search~\citep{shrivastava2014asymmetric} can be utilized. The reader, on the other hand, leverages the freedom of the \emph{one-tower} architecture (Figure~\ref{fig:teaser}a), which takes both the question and the document as a concatenated input and is allowed to process them jointly to obtain a more accurate answer. The reader, however, clearly has a linear-time complexity with respect to the input size.

It is hence commonly conceived that the main role of the retriever is the gain of efficiency at the cost of accuracy. Theoretically, this makes sense; the two-tower architecture enforces the information in the question or the document to be bottlenecked by their embeddings, which can cause the loss of information, and furthermore, they only interact through similarity (metric or inner product) space, which further limits its capability. This especially corresponds well with the motivation for the kernel method in SVMs~\citep{cortes1995support}, where one would need an infinite number of dimensions for the feature map (two-tower) to exactly mimic even a simple kernel (one-tower) such as an RBF kernel~\citep{RBF} in inner product space. After all, any target function that a two-tower model can learn is clearly also learnable by a one-tower model.

Nevertheless, we find some empirical hints from the previous works that somewhat contradict this general belief. For instance, \citet{docqa} and \citet{lewis2020retrieval} both demonstrate in Figure 3 that as the reader reads more top-k documents, the end-to-end QA accuracy somewhat decays. While these observations were not seriously discussed previously, they are clearly worth a closer look because they imply that their retrievers are not just reducing the search space for efficiency, but also playing some role in making the reader more accurate. Is it simply because the readers were just not trained properly?

In this paper, we delve into these observations and find that that the retriever is not merely an approximator of the reader for the document ranking purpose.
In the first part (Section~\ref{sec:strict}), we empirically show that the two-tower model (retrieval-based approach) is not only efficient but also essential for creating a good open-domain question answering model.
That is, the retriever and the reader are complementary to each other, where each has a comparative advantage over the other for accuracy.
We guess that the architectural constraint of the retriever, which might have been originally intended for an approximation, seems to also make the model more robust against various negative examples in large-scale search.
Then in the second part (Section~\ref{sec:model}), we propose a distillation method to enhance the retriever so that the retriever can absorb the strength of the reader while keeping its comparative benefit. Our method is able to enhance the recall rate and the end-to-end QA accuracy of off-the-shelf retrievers~\citep{karpukhin2020dense}, especially with a significant improvement in top-1 recall and accuracy.\footnote{We will make all of our code and model weights publicly available.}

\section{Related Work}\label{sec:related}
\paragraph{Open-Domain QA} Answering open-domain factoid questions can utilize either a highly structured knowledge source (e.g., Knowledge Graph) or large unstructured data (e.g., Wikipedia). In this paper, we primarily focus on the latter, and more specifically ``Retrieve \& Read'' paradigm~\citep{chen2017reading, guu2020realm}, which has been steadily predominant in the community and yielding both high accuracy and efficiency. Note that the act of ``reading'' can be either extractive (i.e., the finding where the answer phrase lies in the documents)~\citep{karpukhin2020dense} or generative~\citep{wang2016machine, lewis2020retrieval, izacard2020leveraging}. More recently, ``closed-book'' QA models have been drawing attention~\citep{roberts2020much, brown2020language}, where all knowledge is encoded in the parameters of the model, and no external knowledge source is attached (i.e., fully parametric).
These models are however also known for being computationally expensive since they require many orders of more parameters to \emph{memorize} the knowledge corpus, and the answers are often unreliable and uninterpretable.
Another line of recent work~\citep{seo2019real} makes the task as a pure retrieval problem by indexing all answer phrase candidates in the knowledge corpus, which makes the inference more efficient and end-to-end. 
Nevertheless, these models can only yield only extractive answers (than generative) and require much larger storage to keep the index.

\paragraph{Document Retrieval \& Ranking} Retrieving relevant documents given a query has been a significant interest in both research and industry, due to its direct applicability to search engines and question answering systems.
It has been known in the community that directly providing the retrieved documents in the order of their retrieval scores often results in poor search quality~\citep{lee1997document}. 
Therefore, in practice, most search engines utilize a ranker that looks into the retrieved top-k documents and reorders them~\citep{rerank, BERTSearch}.
Unlike the retriever, the ranker often employs the one-tower architecture to fully prioritize the search quality over efficiency.
From the QA perspective, ranking can be considered to be either explicitly~\citep{karpukhin2020dense} or implicitly~\citep{lewis2020retrieval} embodied in its reader component.

\paragraph{Similarity Search} In order to find a relevant item (e.g., document) in an extremely large search space, the items and the query are often embedded into a common vector space (using two-tower architecture), and then an off-the-shelf similarity search library such as \texttt{faiss}~\citep{8733051}\footnote{https://github.com/facebookresearch/faiss} can be used for fast retrieval.
The search process can be especially more efficient when an approximation is used.
Traditionally, metric space such as cosine distance or L2 has been a popular choice for the similarity function, which allows us to use approximation methods such as Locality Sensitive Hashing (LSH)~\citep{gionis1999similarity} or k-means clustering~\citep{hartigan1979algorithm} by leveraging the nice theoretical properties of metric space.
However, for many recent question answering models, inner product space seems to lead to a similar or better model than L2 during training~\citep{seo2019real,karpukhin2020dense}, where one can still utilize asymmetric LSH (aLSH)~\citep{shrivastava2014asymmetric} or k-means clustering.
We also adopt inner product space with k-means clustering.

\section{Is Retriever Merely an Approximator of Reader?}\label{sec:strict}
In Section~\ref{subsec:formulation}, we first formally define the open-domain question answering task, in order to formulate our question of interest in a formal manner as well. Then in Section~\ref{subsec:case} and~\ref{subsec:env}, we provide several empirical explanations for the relationship between the retriever and the reader. 

\subsection{Problem Formulation}\label{subsec:formulation}

The task of open-domain QA is to learn a \emph{reader} function $\bm{f}$ that maps two inputs, question $\bm{q}$, and knowledge data $\bm{d}$, to the  answer $\bm{\hat{a}}$, i.e., $\bm{\hat{a}} = \bm{f}(\bm{q}, \bm{d})$.
Note that in the case of closed-book question answering (as discussed in Section~\ref{sec:related}), the knowledge corpus is only observed during training, so the function only takes the question as the input. Here, we will focus on the more typical open-book case, where the identical knowledge corpus is observed during both training and inference.

The classic challenge of the task is that the knowledge corpus $\bm{d}$ is too big, 
so applying the linear-time function $\bm{f}$ on the entire $\bm{d}$ is intractable.
Therefore, a common practice is to create a \emph{retriever} function $\bm{g}$, which extracts a subset of the knowledge corpus in sublinear-time, that would be small enough for efficiency and has sufficient information to infer the answer to the question. 
The retriever usually adopts two-tower architecture (Figure~\ref{fig:teaser}b), whose structural constraint allows us to efficiently perform the subset extraction via approximate maximum inner product search (or nearest neighbor search). More concretely, the subset $\bm{d}'\subset \bm{d}$ (assuming $\bm{d}$ consists of $N$ chunks such as documents, i.e., $\bm{d} = \{\bm{d}_1, \dots, \bm{d}_N\}$ for convenience) is obtained by

\begin{equation}\label{eqn:retriever}
\bm{d}' = \bm{g}(\bm{q}, \bm{d}) =  \text{top-k}_{\bm{d}_i} \psi(\bm{q}) \cdot \phi(\bm{d}_i)
\end{equation}
where $k$ is the target number of chunks, $\psi, \phi$ are feature map functions that map the question and each knowledge chunk to a common $d$-dimensional vector space, respectively,  and $\cdot$ is an inner product operator (we can replace it with L2-distance if we are using nearest neighbor search).

We let $\bm{f}'$ represent the resulting combined model, i.e., $\bm{\hat{a}} = \bm{f}(\bm{q}, \bm{d}) \approx  \bm{f}(\bm{q}, \bm{g}(\bm{q}, \bm{d})) = \bm{f}'(\bm{q}, \bm{d})$. Then we can consider $\bm{f}'$ as an approximator of $\bm{f}$, which gives us efficiency benefit, possibly at the cost of accuracy.
Indeed, an important limitation of $\bm{f}' $ is that the introduction of $\bm{g}$ is a strong structural constraint, which means it is not able to model some relationships that $\bm{f}$ can easily do. An easy example is an RBF Kernel~\citep{RBF}, which can be easily modeled using a one-tower model but cannot be modeled in inner product space since it requires an infinite number of dimensions. Apparently, $\bm{f}$ is strictly more expressible than $\bm{f}'$, so the theoretical answer to whether $\bm{f}'$ has any accuracy (not efficiency) benefit compared to $\bm{f}$ would be no, and we would conclude that $\bm{f}'$ is \emph{merely an approximator}.
\footnote{That is, to be precise, we are comparing between the retriever-augmented reader and the pure reader.}

However, in practice, the structural constraint in $\bm{g}$ induces a representational bottleneck that seems to benefit $\bm{f}'$ accuracy-wise during training. That is, we think $\bm{f}'$ is \emph{not merely an approximator} of $\bm{f}$. We will carefully explore this claim in this section. 
In Section~\ref{subsec:case}, we first perform a case study on a popular reader and retriever model in the literature~\citep{karpukhin2020dense} and observe that the results agree with our hypothesis. We however note that there could be possibly other factors that may have affected this observation, so we adjust the training environment for a fair comparison in Section~\ref{subsec:env}, and again observe the empirical evidence for our hypothesis.

\subsection{Case Study on DPR}\label{subsec:case}
We first start with a simple case study on the off-the-shelf retriever and reader (DPR) by~\citet{karpukhin2020dense}.
An ideal, full-scale experiment would be to compare the reader's accuracy with and without the retriever. However, reading the entire knowledge corpus (e.g., Wikipedia) will take too much time, given limited computational resources.
We instead indirectly compare them on 1,000 questions randomly sampled from the test set of NaturalQuestions (NQ)~\citep{47761}, by varying the number of retrieved documents for each question ($k$) from 1 to 2,000 (a large but much more manageable number than the total number of documents) and analyze the reader's accuracy with the retrieved documents. 
When $k=1$, the influence of the retriever on the reader's accuracy is maximized, as the correct document must be retrieved for the reader to get the answer right.
As $k$ increases, the influence of the retriever decreases, and the reader gets more room for improvement at the cost of slower speed.

The red graph in Figure~\ref{fig:em_recall}a shows the accuracy of the reader at $k=1,\cdots,2000$ (in log scale). We observe that its accuracy peaks at $k=30$ and steadily decreases. This agrees with the previous observations by~\citet{docqa} and \citet{lewis2020retrieval} that the reader accuracy peaks at a relatively low value of $k$. 
While not seriously discussed before, this is a rather extraordinary behavior because the reader's architecture (one-tower, Figure~\ref{fig:teaser}a) is inherently more expressive than the retriever's architecture (two-tower, Figure~\ref{fig:teaser}b).
One possible explanation for this discrepancy could be due to the different training setup between the retriever and the reader, where the retriever observes more negative examples (using in-batch negative training), which are also more random (top-k from random).
Therefore, we do experiments below on DPR reader under a training environment similar to that of the retriever.

\subsection{Reader Training under Setting Similar to Retriever's}\label{subsec:env}

\begin{figure}[t] 
\centering
\begin{subfigure}[b]{0.47\textwidth}
\centering
\caption{End-to-end QA accuracy (Exact Match, y-axis on the left) of DPR reader and the retrieval recall rate (y-axis on the right) of DPR retriever.}
\includegraphics[width=\textwidth]{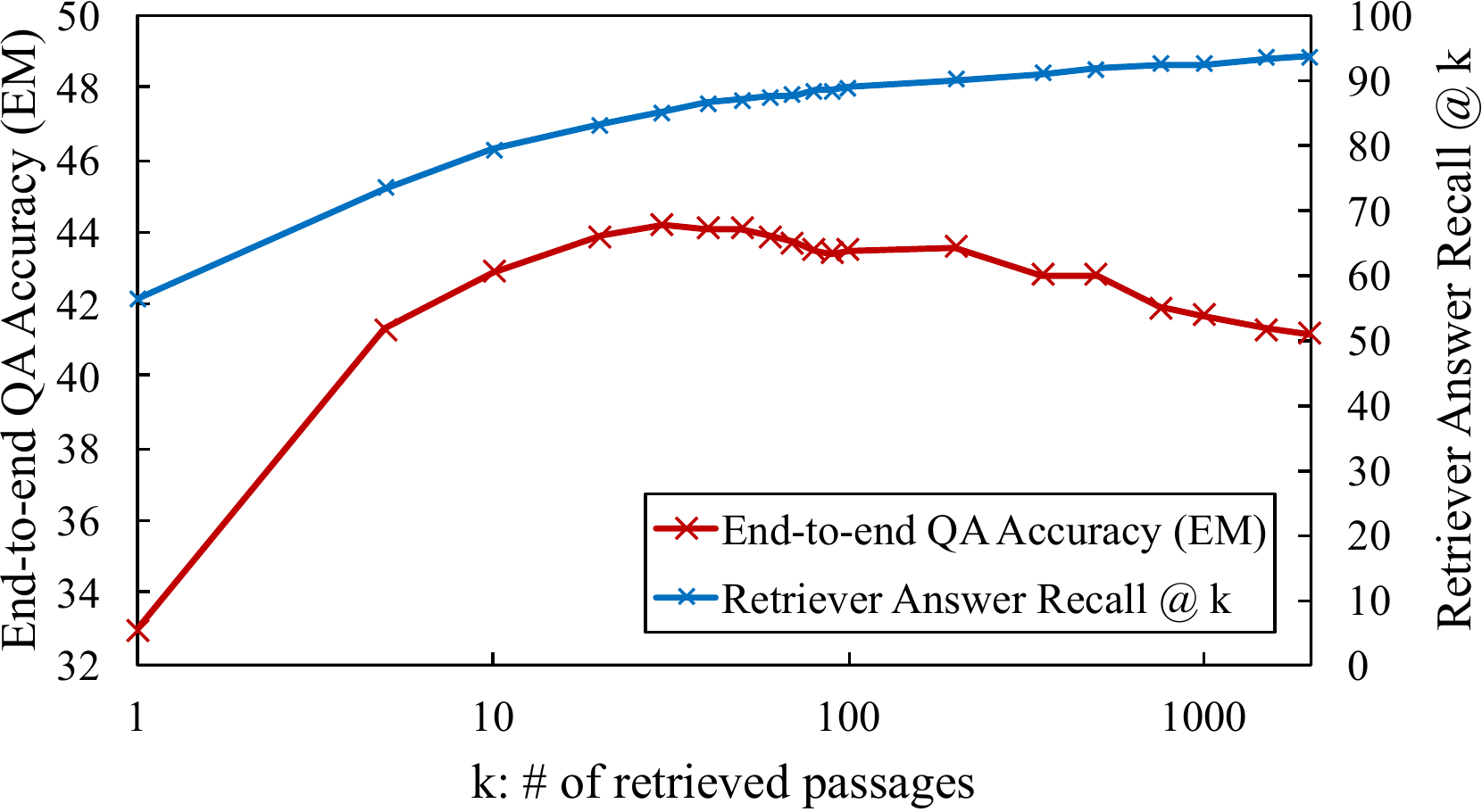}
\end{subfigure}
\hfill
\begin{subfigure}[b]{0.49\textwidth}
\centering
\caption{\small End-to-end QA Accuracy (Exact Match) of DPR readers trained under various training environments.}
\includegraphics[width=\textwidth]{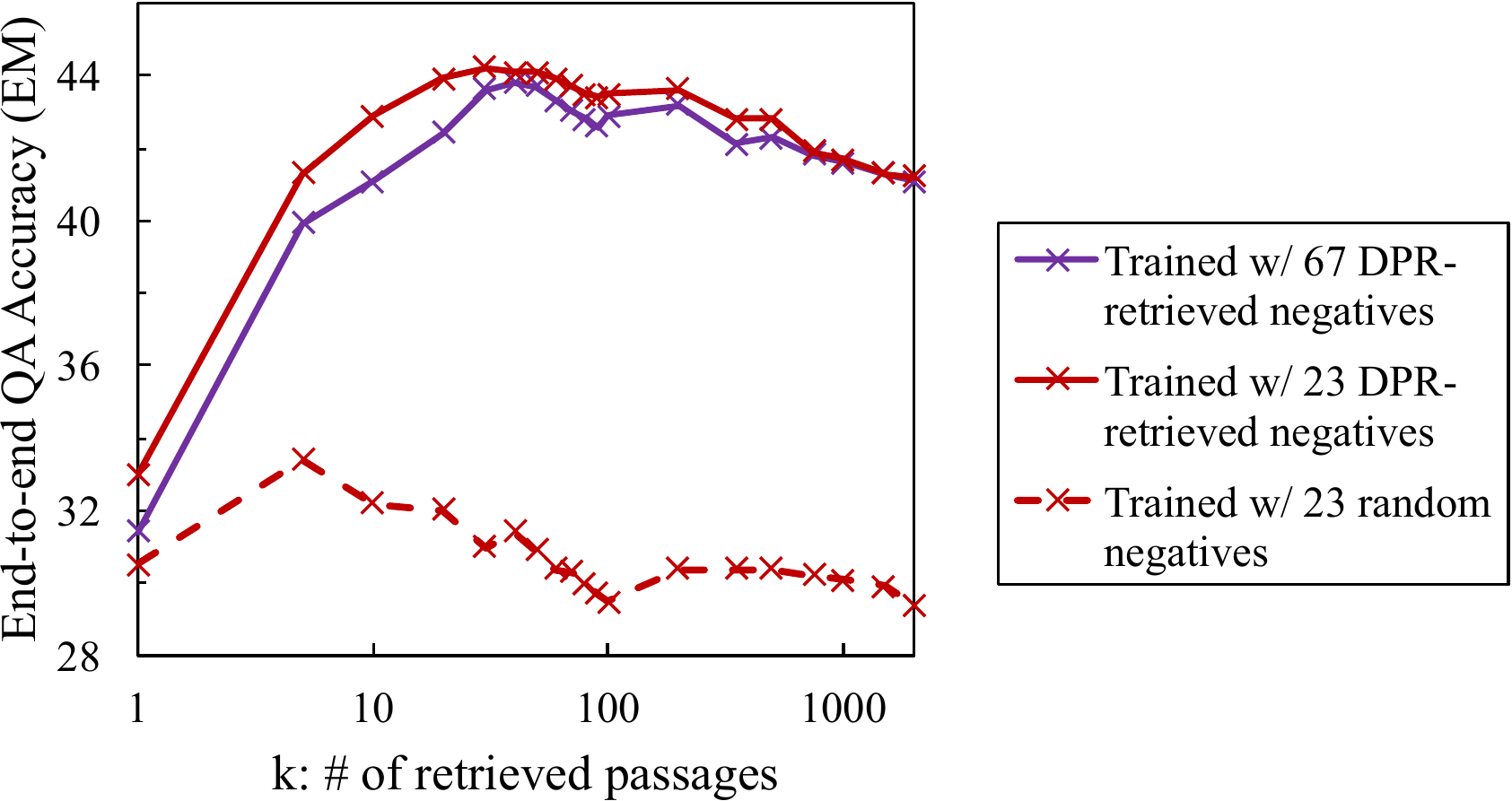}
\end{subfigure}
\caption{\small Results of the experiments using DPR. All experiments are done with 1,000 questions sampled from the test set of NaturalQuestions.}
\label{fig:em_recall}
\end{figure}

Here we investigate whether the performance drop of the reader with increasing $k$ after the peak comes from (1) training with less random negatives or (2) training with smaller number of negative examples.
For the former investigation, we set the number of negatives examples to 23 and examine two sampling strategies; one is where the negative examples are randomly sampled from the entire corpus, and the other is where they are obtained from the top-k documents retrieved by the original DPR retriever. It is hence expected that the former training setup allows the reader to observe more diverse negative examples, while the latter setup allows the reader to observe harder negative examples, as also noted by~\citet{karpukhin2020dense}.
For the latter examination, we fix the sampling strategy as obtaining the negative examples from DPR retriever, and train the reader with different numbers of negative examples: 23 (original DPR reader) and 67. To train the reader with 67 negatives while preventing GPU OOM, we use the batch size of 8 and the gradient accumulation step of 2 (the original DPR reader uses the batch size of 16 for training).\footnote{By nature, the one-tower reader, which jointly models a question-document pair each time, is difficult to be trained with a large batch size or a large number of negative passages, unlike the two-tower retriever.}
We vary $k$ in the same range of 1 to 2,000 and report the reader's accuracy in a similar manner as in Figure~\ref{fig:em_recall}a.

Figure~\ref{fig:em_recall}b shows the results of both experiments.
From the red graphs, which represent the result of the former investigation, we can clearly see that the reader's accuracy decreases as $k$ increases regardless of the sampling strategy, that the solid line (DPR-retrieved negatives) and dashed line (random negatives) peak at small $k$ of 30 and 5, respectively.
Likewise, increasing the number of negatives also does not seem to increase the number $k$ where the peak is achieved; the peak for each of the models trained with 23 and 67 negatives are achieved at similar $k$ of 30 and 40, respectively.
Such results that the reader's accuracy still peaks at a relatively low $k$ even under the training setup similar to that of the retriever seem to indicate that the retriever is indeed giving a non-trivial accuracy benefit to the reader.

\section{Distilling Reader into Retriever}\label{sec:model}

As seen in Section~\ref{sec:strict}, the two-tower retriever and one-tower reader are in a complementary relationship to each other; our guess is that the former is good at finding passage embeddings relevant to a question embedding from a large vector space, and the latter is good at distinguishing between difficult examples in a relatively small search space. Therefore, here we propose a method to combine the best of both worlds to enhance the performance of the retriever: \textit{distilling the knowledge of the reader into the retriever}.

Several learning methods can be applied or explored to transfer the knowledge of the reader to the retriever, but we specifically utilize knowledge distillation~\citep{hinton2015distilling} in this work. Knowledge distillation is often used as a technique for model compression, enabling the generalization ability of a large model to be transferred to a smaller model without much loss in performance~\citep{sanh2019distilbert, liu2019improving}. On the other hand, here we employ knowledge distillation to transfer the representational knowledge learned by the expressive one-tower reader to the two-tower retriever, which has a relatively bottlenecked architecture.

Hereafter, we describe how the two-tower retriever of DPR~\citep{karpukhin2020dense}, which is actively adopted in the latest open-domain question answering models to achieve state-of-the-art results~\citep{lewis2020retrieval, izacard2020leveraging}, can be enhanced by distillation from the one-tower reader proposed in the same work. Although this paper presents one example of utilizing distillation, the approach itself is simple and intended to be generic, not strictly tied to any particular retriever or reader, as long as the knowledge of the reader can be distilled into the retriever in any form.

\subsection{Distillation Setup}

\paragraph{Student: Retriever}
Following the architecture of DPR retriever, our proposed \Ours{} (\ours{}) is a two-tower model that consists of two dense encoders: $E_Q(\cdot)$, which maps a question $q$ to a vector space, and $E_D(\cdot)$, which maps each knowledge chunk $d_i$, a block of 100 words dubbed passage in this work, to a common vector space.

Let us denote the list of passages that the retriever is scoring on behalf of a question $q$ as $D_\text{ret}$. Then, the retrieval score $\text{sim}_\text{ret}(q, d_i)$ is calculated by the inner product between $q$ and each passage $d_i \in D_\text{ret}$. The top-scoring passages that would serve as the input to the reader, $D_\text{read}$, are inferred as follows:

\begin{equation}
\begin{aligned}
    & \text{sim}_\text{ret}(q, d_i) = E_Q(q) \cdot E_D(d_i), \\
    & D_\text{read} = \big\{d_k \mid k \in \operatorname*{argmax}_i\ \text{sim}(q, d_i), \forall d_i \in D_\text{ret} \big\}.
\end{aligned}
\end{equation}

\paragraph{Teacher: Reader}
DPR reader described in Section 6.1 of the work of \citet{karpukhin2020dense} is the teacher of our retriever. As input, the model takes the concatenation of the token embeddings of question $q$, the title, and contents of each passage $d_i$ in $D_\text{read}$. Along with the scores of each token being the starting or ending positions of the answer, the reader outputs a ranking score $\text{sim}_\text{read}(q, d_i)$ for each passage $d_i \in D_\text{read}$.

\paragraph{Objective Function}

Let each of $z_\text{ret}$ and $z_\text{read}$ be a $|D_\text{read}|$-dimensional vector, where each element is the score for a question-passage pair in the inputs to the reader, $(q, d_i), \forall d_i \in D_\text{read}$. For model $\in$ \{ret, read\},

\begin{equation}
    z_\text{model} = \big[ \text{sim}_\text{model}(q, d_1), \dots, \text{sim}_\text{model}(q, d_{|D_\text{read}|}) \big].
\end{equation}

To turn the score vectors $z_\text{ret}$ and $z_\text{read}$ into probability distributions without saturation, softmax with temperature $T$ is applied. We then minimize $D_{KL}(P_\text{read} || P_\text{ret})$, where each of the probability distributions is calculated as follows:

\begin{equation}
    P_\text{model} = \bigg[
    \frac{
        \exp (z_{{\text{model}}_1} / T)
        } {\sum_j \exp (z_{{\text{model}}_j} / T)
    },
    \cdots,
    \frac{
        \exp (z_{{\text{model}}_{|D_\text{read}|}} / T)
        } {\sum_j \exp (z_{{\text{model}}_j} / T)
    }
    \bigg].
\end{equation}

\subsection{Baseline Models and Datasets}
We choose DPR~\citep{karpukhin2020dense} as the main baseline to apply our proposed method to enhance retrieval performance. We run all the retrieval experiments on top of its official implementation, model weights, and evaluation scripts.\footnote{\href{https://github.com/facebookresearch/DPR}{https://github.com/facebookresearch/DPR}} We also investigate the change in the end-to-end QA accuracy when RDR is used together with the readers of DPR and RAG~\citep{lewis2020retrieval}. The experiments related to RAG made use of the implementation, model weights, \texttt{FAISS} index, and evaluation scripts recently made public in huggingface/transformers~\citep{Wolf2019HuggingFacesTS}.\footnote{\href{https://github.com/huggingface/transformers}{https://github.com/huggingface/transformers}} We consider NaturalQuestions~\citep{47761} as our main dataset because both DPR and RAG officially provides only the weights and indices for NaturalQuestions. Although we successfully reproduced and thus could report the results of experiments using DPR for TriviaQA~\citep{joshi2017triviaqa}, we could not test the other settings due to the lack of training resources, e.g., size of RAM and number of available GPUs.

\subsection{Training Details}\label{subsec:training}

\paragraph{Retriever}
To enhance DPR retriever to get \ours{}, we initialize the model with the pretrained DPR retriever and finetune it via knowledge distillation from DPR reader. The architecture of the retriever is thus the same with that of DPR, which consists of two bert-base encoders with the hidden dimension of 768. We set the knowledge distillation temperature to 3 and the rate of distillation loss to 1.0. We use AdamW optimizer with an initial learning rate of 1e-5 and warmup steps of 100. The other hyperparameters mostly follow the setup used to train DPR reader. We train \ours{} for 16 epochs with a batch size of 10 and number of passages of 16, and select the checkpoint which reports the highest retrieval accuracy on 10\% or 100\% of the validation set, where the former is used as an option to shorten the training time for some experiments.
More details are in Appendix~\ref{appendix:tech}.

\paragraph{Reader}
In order to see whether \ours{} can lead to the enhancement of not only retrieval recall rate but also end-to-end open-domain QA accuracy, we perform more experiments using the pretrained readers of DPR and RAG. As discussed in Section~\ref{subsec:em}, the readers of DPR and RAG are dependent on the original retrievers they used at training time, so just replacing the retrievers with RDR during inference creates a discrepancy in the input distribution between the training and inference time. We hence finetune the reader for 3 to 6 epochs and select the model with the best Exact Match (EM) scorer on 10\% or 100\% of the validation set. We use the same hyperparameters to finetune DPR reader and RAG reader, except that we set the learning rate to 1e-5 for the latter and batch size to 4 to meet the budget of training resources.

\section{Experimental Results}

\subsection{Retrieval Recall Rate}

\begin{table}[ht!]
    \setlength{\tabcolsep}{3.5pt}
    \centering
    \caption{Retrieval recall rate of DPR~\citep{karpukhin2020dense}, RAG~\citep{lewis2020retrieval}, and RDR (Ours) on NQ-dev, NQ-test, and TriviaQA-test. $\drsh$ indicates which model RDR targets to enhance. $\dagger$ is from \citet{karpukhin2020dense}, and $\ddagger$ is approximated from Figure 3 of \citet{lewis2020retrieval}.}
    \label{table:main_recall}
    \footnotesize
    \begin{threeparttable}
    \begin{tabular*}{1.0\textwidth}{l@{\extracolsep{\fill}}|cccc|cccc|cccc}
        \toprule
                          \textbf{Dataset} & \multicolumn{4}{c|}{\textbf{NQ-dev}}  & \multicolumn{4}{c|}{\textbf{NQ-test}}  & \multicolumn{4}{c}{\textbf{TriviaQA-test}}  \\
\midrule
Top-k       & 1      & 20     & 50     & 100  & 1       & 20     & 50     & 100    & 1             & 20     & 50     & 100    \\
\midrule
\midrule
DPR-Single\;             & 44.2$^\ddagger$   & 76.9$^\ddagger$   & 81.3$^\ddagger$   & 84.2 & 46.3    & 78.4$^\dagger$   & 84.1   & 85.4$^\dagger$   & 54.4          & 79.4$^\dagger$   & 82.9   & 85.0$^\dagger$   \\
$\drsh$ w/ RDR & \textbf{54.1}   & \textbf{80.7}   & \textbf{84.1}   & \textbf{85.8}        & \textbf{54.2}    & \textbf{82.8}   & \textbf{86.3}   & \textbf{88.2}   & \textbf{62.5} & \textbf{82.5}   & \textbf{85.7}   & \textbf{87.3}   \\
 & (+9.9) & (+3.8) & (+2.8) & (+1.6)   & (+7.9)  & (+4.4) & (+2.2) & (+2.8) & (+8.1)        & (+3.1) & (+2.8) & (+2.3) \\
\midrule
SOTA            & 51.7$^\ddagger$   & 79.2$^\ddagger$   & 83.0$^\ddagger$    & -    & -       & 79.4$^\dagger$   & -      & 86.0$^\dagger$   & -             & 79.9$^\dagger$   & -      & 85.0$^\dagger$   \\
        \bottomrule
    \end{tabular*}
    \end{threeparttable}
\end{table}

Table~\ref{table:main_recall} shows the retrieval recall rate of several retrievers, measured as the percentage of the passages that contain the answer among the top-k retrieved ones, on the dev and test set of NaturalQuestions (NQ) and test set of TriviaQA. The compared models are  DPR~\citep{karpukhin2020dense}, RAG~\citep{lewis2020retrieval}, and RDR (Ours). RDR is built on top of DPR trained with a single dataset, so we use an arrow mark to indicate that it specifically improves DPR-Single, not DPR-Multi or BM25+DPR.

\begin{figure}[htb]
\centering
\begin{minipage}[c]{0.52\textwidth}
    \centering
    \includegraphics[width=1.0\textwidth]{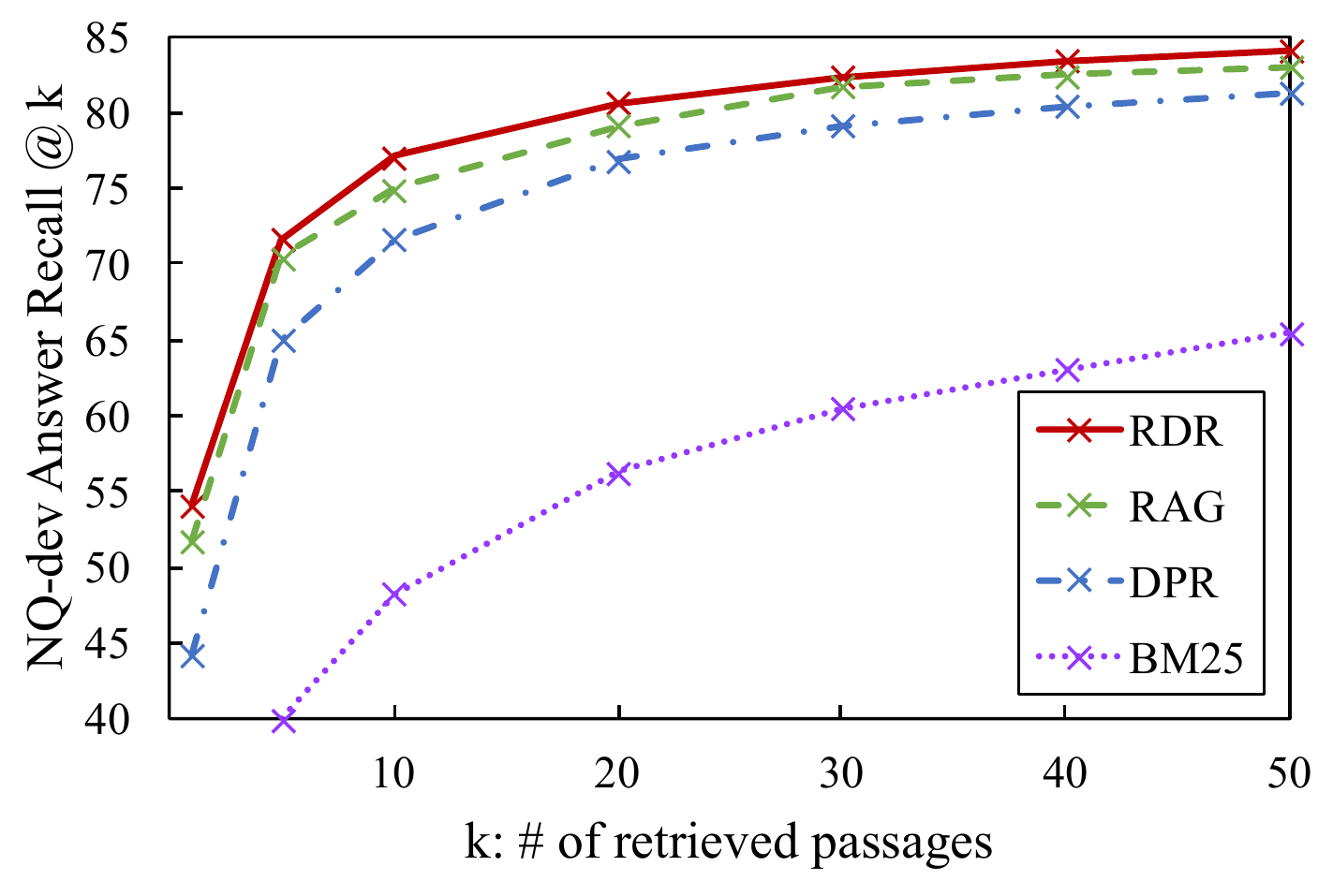}
    \caption{\small Answer recall rate of several models on the dev set of NQ measured as the fraction of the passages that contain the answer among the retrieved top-k ones. The values for the models other than RDR are approximated from Figure 3 of the work of \citet{lewis2020retrieval}.}
    \label{fig:recall_graph}
\end{minipage}
\hfill
\begin{minipage}[c]{0.45\textwidth}
    \captionof{table}[]{\small The ablation study result, which shows how the retrieval recall rate on the test set of NaturalQuestions changes when RDR enhanced from DPR-Single is trained without distillation. The recall rate shows a consistent drop at all $k$'s when no distillation is used, and the gap is relatively large at top-1.}
    \label{table:ablation_recall}
    \footnotesize
    \begin{threeparttable}
    \begin{tabular*}{1.0\textwidth}{l@{\extracolsep{\fill}}cc}
\toprule
                     & DPR-Single w/ RDR    & w/o distillation \\
\midrule
Top-1                     & 54.2                    & 52.4 (-1.8)      \\
Top-20                    & 82.8                    & 82.6 (-0.2)      \\
Top-50                    & 86.3                    & 85.7 (-0.6)      \\
Top-100                   & 88.2                    & 87.5 (-0.7)   \\
        \bottomrule
    \end{tabular*}
    \end{threeparttable}
\end{minipage}
\end{figure}

To compare RDR with the state-of-the-art retrievers, we present the best results of the previous works for each dataset in the last row of the table. The results with $\dagger$ and $\ddagger$ are borrowed from Table 2 of the work of \citet{karpukhin2020dense} and approximated from Figure 3 of the work of \citet{lewis2020retrieval}, respectively. Following Figure 3 of \citet{lewis2020retrieval}, we also show the recall of several models on the dev set of NatrualQuestions in Figure~\ref{fig:recall_graph}.

As shown in the table, RDR consistently outperforms DPR-Single by a wide margin and shows state-of-the-art retrieval performance throughout all datasets and the number of retrieved passages. The performance gap between RDR and DPR-Single is especially large at top-1 and when $k$ is smaller.
The significant improvement in the retrieval recall rate at small k's is especially beneficial to end-to-end open-domain QA, because it opens up more possibility to the reader to get the answer right while seeing fewer passages, which is often important for answering user's question in real-time.\footnote{How the latency increases with respect to the number of passages read at inference time is reported in Appendix~\ref{appendix:latency}.}

\subsection{End-to-end Open-Domain Question Answering Accuracy}\label{subsec:em}

\begin{table}[ht!]
    \setlength{\tabcolsep}{3.5pt}
    \centering
    \caption{\small Enhancement in end-to-end QA accuracy on NaturalQuestions and TriviaQA achieved by utilizing RDR along with the readers of DPR-Single and RAG-Token. We finetune the readers for a few epochs. The values in the ``Reported'' column for DPR-Single-related models are the best performance achieved among $k \in \{1, 10, 20, 30, 40, 50\}$, whereas those values for RAG-Token-related models are inferred with $k=15$, and the way to choose $k$ follows the original inference setup of the baseline models. We could not perform RAG-related experiments on TriviaQA because the model checkpoint is not publicly available, and it was non-trivial to reproduce the with our limited computation resources.}
    \label{table:main_em}
    \footnotesize
    \begin{threeparttable}
\begin{tabular*}{1.0\textwidth}{l@{\extracolsep{\fill}}cccccc}
\toprule
\textbf{Dataset} & \multicolumn{3}{c}{\textbf{NQ-test}}  & \multicolumn{3}{c}{\textbf{TriviaQA-test}} \\
\cmidrule{2-4} \cmidrule{5-7}
              & Top-1       & \multicolumn{2}{c}{Reported}  & Top-1         & \multicolumn{2}{c}{Reported} \\
\cmidrule{2-2} \cmidrule{3-4} \cmidrule{5-5} \cmidrule{6-7}
                      & EM          & EM          & Top-k  & EM            & EM          & Top-k  \\
\midrule
\midrule
DPR-Single            & 32.3        & 41.5        & 50 & 44.5          & 56.8        & 50 \\
$\drsh$ w/ RDR & 37.3 (+5.0) & 42.1 (+0.6) & 10 & 49.1 (+4.6)          & 57.0 (+0.2) & 50 \\
\midrule
RAG-Token             & 39.4        & 44.1        & 15 & -             & 55.2        & -  \\
$\drsh$ w/ RDR & 40.9 (+1.5) & 44.5 (+0.4) & 15 & -             & -           & -  \\
        \bottomrule
    \end{tabular*}
    \end{threeparttable}
\end{table}

In order to see if the performance gain in passage retrieval can also lead to the enhancement in end-to-end open-domain QA accuracy, we replace the retrievers of DPR-Single and RAG-Token with RDR and measure the change in the Exact Match (EM) score.  However, since the readers are trained with the outputs of their original retrievers, just replacing the retrievers with RDR at the inference time creates a gap in the input distribution between the training and inference phases. Therefore, we finetune the readers for a few epochs and report the result in Table~\ref{table:main_em}.

The consistent gain in the EM score obtained with RDR suggests that the enhanced retrieval performance can also improve the end-to-end open-domain QA accuracy. The large gain in EM at reading only one passage (Top-1) would have come from the significant improvement in the top-1 retrieval recall rate.
On the other hand, the improvement in the end-to-end accuracy is not proportional to the increase in retrieval performance; the gain in the former is relatively small. Our assumption is that just finetuning the reader with respect to the retriever may not be sufficient for the reader to fully benefit from the enhanced retriever, and we leave the investigation of more effective learning strategies for the reader as future work.
See Appendix~\ref{appendix:sota} for other state-of-the-art QA models.

\paragraph{Ablation Studies}

Table~\ref{table:ablation_recall} reports the result when the retriever is trained without distillation. There is a consistent drop in the retrieval recall rate when distillation is not used, and the gap is relatively large at Top-1.
Regarding distillation, we also tried training RDR with temperature values of $T \in \{0.5, 1, 2, 3, 4\}$ and observed from the retrieval accuracy graph on NQ-dev during training that the area under the trend line is wider with larger T.\footnote{There was no significant difference in the retrieval recall rate between the models trained with $T=3$ and $T=4$, so we chose $T=3$ for all other \ours{} experiments.}
Appendix~\ref{appendix:ablations} shows the ablation results where the reader is not finetuned with respect to RDR, and in general, there is a drop in EM without finetuning, which may have come from the input distribution gap between the training and inference phase.

\section{Conclusion}
In this paper, we revisit a rather overlooked question in open-domain question answering (QA) of whether the two-tower architecture (retriever) can be entirely replaced by the one-tower architecture (reader) for accuracy if unlimited computation resources and time are allowed. 
Our empirical evidence indicates that the answer is no, and that the presence of the retriever seems to be essential for not only the efficiency but also the accuracy of a QA model, presumably because it becomes more robust against diverse negative documents.
In the second part, given that the retriever and the reader are complementary to each other, we propose to distill the reader into the retriever to combine the best of both worlds.
Our distillation method significantly enhances the recall rate of an existing retriever, which also translates into a non-trivial improvement in the end-to-end QA accuracy. 
Future work includes scaling up the experiment so that our method can also be evaluated on more recent models that require a large number of (e.g. 50+) GPUs to train.

\bibliography{main}

\begin{thebibliography}{32}
\providecommand{\natexlab}[1]{#1}
\providecommand{\url}[1]{\texttt{#1}}
\expandafter\ifx\csname urlstyle\endcsname\relax
  \providecommand{\doi}[1]{doi: #1}\else
  \providecommand{\doi}{doi: \begingroup \urlstyle{rm}\Url}\fi

\bibitem[Asai et~al.(2020)Asai, Hashimoto, Hajishirzi, Socher, and
  Xiong]{asai2019learning}
Akari Asai, Kazuma Hashimoto, Hannaneh Hajishirzi, Richard Socher, and Caiming
  Xiong.
\newblock Learning to retrieve reasoning paths over wikipedia graph for
  question answering.
\newblock In \emph{ICLR}, 2020.

\bibitem[Bachrach et~al.(2014)Bachrach, Finkelstein, Gilad-Bachrach, Katzir,
  Koenigstein, Nice, and Paquet]{bachrach2014speeding}
Yoram Bachrach, Yehuda Finkelstein, Ran Gilad-Bachrach, Liran Katzir, Noam
  Koenigstein, Nir Nice, and Ulrich Paquet.
\newblock Speeding up the xbox recommender system using a euclidean
  transformation for inner-product spaces.
\newblock In \emph{Proceedings of the 8th ACM Conference on Recommender
  systems}, pp.\  257--264, 2014.

\bibitem[Brown et~al.(2020)Brown, Mann, Ryder, Subbiah, Kaplan, Dhariwal,
  Neelakantan, Shyam, Sastry, Askell, et~al.]{brown2020language}
Tom~B Brown, Benjamin Mann, Nick Ryder, Melanie Subbiah, Jared Kaplan, Prafulla
  Dhariwal, Arvind Neelakantan, Pranav Shyam, Girish Sastry, Amanda Askell,
  et~al.
\newblock Language models are few-shot learners.
\newblock \emph{arXiv preprint arXiv:2005.14165}, 2020.

\bibitem[Carbonell \& Goldstein(1998)Carbonell and Goldstein]{rerank}
Jaime Carbonell and Jade Goldstein.
\newblock The use of mmr, diversity-based reranking for reordering documents
  and producing summaries.
\newblock In \emph{SIGIR}, 1998.

\bibitem[Chang et~al.(2010)Chang, Hsieh, Chang, Ringgaard, and Lin]{RBF}
Yin-Wen Chang, Cho-Jui Hsieh, Kai-Wei Chang, Michael Ringgaard, and Chih-Jen
  Lin.
\newblock Training and testing low-degree polynomial data mappings via linear
  svm.
\newblock \emph{JMLR}, 11\penalty0 (4), 2010.

\bibitem[Chen et~al.(2017)Chen, Fisch, Weston, and Bordes]{chen2017reading}
Danqi Chen, Adam Fisch, Jason Weston, and Antoine Bordes.
\newblock Reading wikipedia to answer open-domain questions.
\newblock In \emph{ACL}, 2017.

\bibitem[Clark \& Gardner(2017)Clark and Gardner]{docqa}
Christopher Clark and Matt Gardner.
\newblock Simple and effective multi-paragraph reading comprehension.
\newblock \emph{arXiv preprint arXiv:1710.10723}, 2017.

\bibitem[Cortes \& Vapnik(1995)Cortes and Vapnik]{cortes1995support}
Corinna Cortes and Vladimir Vapnik.
\newblock Support-vector networks.
\newblock \emph{Machine learning}, 20\penalty0 (3):\penalty0 273--297, 1995.

\bibitem[Devlin et~al.(2019)Devlin, Chang, Lee, and Toutanova]{bert}
Jacob Devlin, Ming-Wei Chang, Kenton Lee, and Kristina Toutanova.
\newblock Bert: Pre-training of deep bidirectional transformers for language
  understanding.
\newblock In \emph{NAACL}, 2019.

\bibitem[Gionis et~al.(1999)Gionis, Indyk, Motwani,
  et~al.]{gionis1999similarity}
Aristides Gionis, Piotr Indyk, Rajeev Motwani, et~al.
\newblock Similarity search in high dimensions via hashing.
\newblock In \emph{VLDB}, 1999.

\bibitem[Guu et~al.(2019)Guu, Lee, Tung, Pasupat, and Chang]{guu2020realm}
Kelvin Guu, Kenton Lee, Zora Tung, Panupong Pasupat, and Ming-Wei Chang.
\newblock Realm: Retrieval-augmented language model pre-training.
\newblock In \emph{ICML}, 2019.

\bibitem[Hartigan \& Wong(1979)Hartigan and Wong]{hartigan1979algorithm}
John~A Hartigan and Manchek~A Wong.
\newblock Algorithm as 136: A k-means clustering algorithm.
\newblock \emph{Journal of the royal statistical society. series c (applied
  statistics)}, 28\penalty0 (1):\penalty0 100--108, 1979.

\bibitem[Hinton et~al.(2014)Hinton, Vinyals, and Dean]{hinton2015distilling}
Geoffrey Hinton, Oriol Vinyals, and Jeff Dean.
\newblock Distilling the knowledge in a neural network.
\newblock \emph{NIPS Deep Learning Workshop}, 2014.

\bibitem[Izacard \& Grave(2020)Izacard and Grave]{izacard2020leveraging}
Gautier Izacard and Edouard Grave.
\newblock Leveraging passage retrieval with generative models for open domain
  question answering.
\newblock \emph{arXiv preprint arXiv:2007.01282}, 2020.

\bibitem[{Johnson} et~al.(2019){Johnson}, {Douze}, and {Jégou}]{8733051}
J.~{Johnson}, M.~{Douze}, and H.~{Jégou}.
\newblock Billion-scale similarity search with gpus.
\newblock \emph{IEEE Transactions on Big Data}, pp.\  1--1, 2019.

\bibitem[Joshi et~al.(2017)Joshi, Choi, Weld, and
  Zettlemoyer]{joshi2017triviaqa}
Mandar Joshi, Eunsol Choi, Daniel~S Weld, and Luke Zettlemoyer.
\newblock Triviaqa: A large scale distantly supervised challenge dataset for
  reading comprehension.
\newblock In \emph{ACL}, 2017.

\bibitem[Karpukhin et~al.(2020)Karpukhin, O{\u{g}}uz, Min, Wu, Edunov, Chen,
  and Yih]{karpukhin2020dense}
Vladimir Karpukhin, Barlas O{\u{g}}uz, Sewon Min, Ledell Wu, Sergey Edunov,
  Danqi Chen, and Wen-tau Yih.
\newblock Dense passage retrieval for open-domain question answering.
\newblock In \emph{EMNLP}, 2020.

\bibitem[Kwiatkowski et~al.(2019)Kwiatkowski, Palomaki, Redfield, Collins,
  Parikh, Alberti, Epstein, Polosukhin, Kelcey, Devlin, Lee, Toutanova, Jones,
  Chang, Dai, Uszkoreit, Le, and Petrov]{47761}
Tom Kwiatkowski, Jennimaria Palomaki, Olivia Redfield, Michael Collins, Ankur
  Parikh, Chris Alberti, Danielle Epstein, Illia Polosukhin, Matthew Kelcey,
  Jacob Devlin, Kenton Lee, Kristina~N. Toutanova, Llion Jones, Ming-Wei Chang,
  Andrew Dai, Jakob Uszkoreit, Quoc Le, and Slav Petrov.
\newblock Natural questions: a benchmark for question answering research.
\newblock \emph{TACL}, 2019.

\bibitem[Lee et~al.(1997)Lee, Chuang, and Seamons]{lee1997document}
Dik~L Lee, Huei Chuang, and Kent Seamons.
\newblock Document ranking and the vector-space model.
\newblock \emph{IEEE software}, 14\penalty0 (2):\penalty0 67--75, 1997.

\bibitem[Lee et~al.(2019)Lee, Chang, and Toutanova]{lee2019latent}
Kenton Lee, Ming-Wei Chang, and Kristina Toutanova.
\newblock Latent retrieval for weakly supervised open domain question
  answering.
\newblock In \emph{ACL}, 2019.

\bibitem[Lewis et~al.(2020)Lewis, Perez, Piktus, Petroni, Karpukhin, Goyal,
  K{\"u}ttler, Lewis, Yih, Rockt{\"a}schel, et~al.]{lewis2020retrieval}
Patrick Lewis, Ethan Perez, Aleksandara Piktus, Fabio Petroni, Vladimir
  Karpukhin, Naman Goyal, Heinrich K{\"u}ttler, Mike Lewis, Wen-tau Yih, Tim
  Rockt{\"a}schel, et~al.
\newblock Retrieval-augmented generation for knowledge-intensive nlp tasks.
\newblock In \emph{EMNLP}, 2020.

\bibitem[Liu et~al.(2019)Liu, He, Chen, and Gao]{liu2019improving}
Xiaodong Liu, Pengcheng He, Weizhu Chen, and Jianfeng Gao.
\newblock Improving multi-task deep neural networks via knowledge distillation
  for natural language understanding.
\newblock \emph{arXiv preprint arXiv:1904.09482}, 2019.

\bibitem[Min et~al.(2019{\natexlab{a}})Min, Chen, Hajishirzi, and
  Zettlemoyer]{min2019discrete}
Sewon Min, Danqi Chen, Hannaneh Hajishirzi, and Luke Zettlemoyer.
\newblock A discrete hard em approach for weakly supervised question answering.
\newblock In \emph{EMNLP}, 2019{\natexlab{a}}.

\bibitem[Min et~al.(2019{\natexlab{b}})Min, Chen, Zettlemoyer, and
  Hajishirzi]{min2019knowledge}
Sewon Min, Danqi Chen, Luke Zettlemoyer, and Hannaneh Hajishirzi.
\newblock Knowledge guided text retrieval and reading for open domain question
  answering.
\newblock \emph{arXiv preprint arXiv:1911.03868}, 2019{\natexlab{b}}.

\bibitem[Min et~al.(2020)Min, Michael, Hajishirzi, and
  Zettlemoyer]{min2020ambigqa}
Sewon Min, Julian Michael, Hannaneh Hajishirzi, and Luke Zettlemoyer.
\newblock {A}mbig{QA}: Answering ambiguous open-domain questions.
\newblock In \emph{EMNLP}, 2020.

\bibitem[Nogueira \& Cho(2019)Nogueira and Cho]{BERTSearch}
Rodrigo Nogueira and Kyunghyun Cho.
\newblock Passage re-ranking with bert.
\newblock \emph{arXiv preprint arXiv:1901.04085}, 2019.

\bibitem[Roberts et~al.(2020)Roberts, Raffel, and Shazeer]{roberts2020much}
Adam Roberts, Colin Raffel, and Noam Shazeer.
\newblock How much knowledge can you pack into the parameters of a language
  model?
\newblock \emph{arXiv preprint arXiv:2002.08910}, 2020.

\bibitem[Sanh et~al.(2019)Sanh, Debut, Chaumond, and Wolf]{sanh2019distilbert}
Victor Sanh, Lysandre Debut, Julien Chaumond, and Thomas Wolf.
\newblock Distilbert, a distilled version of bert: smaller, faster, cheaper and
  lighter.
\newblock In \emph{NeurIPS}, 2019.

\bibitem[Seo et~al.(2019)Seo, Lee, Kwiatkowski, Parikh, Farhadi, and
  Hajishirzi]{seo2019real}
Minjoon Seo, Jinhyuk Lee, Tom Kwiatkowski, Ankur~P Parikh, Ali Farhadi, and
  Hannaneh Hajishirzi.
\newblock Real-time open-domain question answering with dense-sparse phrase
  index.
\newblock In \emph{ACL}, 2019.

\bibitem[Shrivastava \& Li(2014)Shrivastava and Li]{shrivastava2014asymmetric}
Anshumali Shrivastava and Ping Li.
\newblock Asymmetric lsh (alsh) for sublinear time maximum inner product search
  (mips).
\newblock In \emph{NIPS}, 2014.

\bibitem[Wang \& Jiang(2017)Wang and Jiang]{wang2016machine}
Shuohang Wang and Jing Jiang.
\newblock Machine comprehension using match-lstm and answer pointer.
\newblock In \emph{ICLR}, 2017.

\bibitem[Wolf et~al.(2019)Wolf, Debut, Sanh, Chaumond, Delangue, Moi, Cistac,
  Rault, Louf, Funtowicz, Davison, Shleifer, von Platen, Ma, Jernite, Plu, Xu,
  Scao, Gugger, Drame, Lhoest, and Rush]{Wolf2019HuggingFacesTS}
Thomas Wolf, Lysandre Debut, Victor Sanh, Julien Chaumond, Clement Delangue,
  Anthony Moi, Pierric Cistac, Tim Rault, Rémi Louf, Morgan Funtowicz, Joe
  Davison, Sam Shleifer, Patrick von Platen, Clara Ma, Yacine Jernite, Julien
  Plu, Canwen Xu, Teven~Le Scao, Sylvain Gugger, Mariama Drame, Quentin Lhoest,
  and Alexander~M. Rush.
\newblock Huggingface's transformers: State-of-the-art natural language
  processing.
\newblock \emph{ArXiv}, abs/1910.03771, 2019.

\end{thebibliography}
\bibliographystyle{iclr2021_conference}

\newpage

\appendix
\section{Appendix}

\subsection{Technical Details}\label{appendix:tech}

\begin{table}[h]
    \setlength{\tabcolsep}{3.5pt}
    \centering
    \caption{Retrieval recall rate, index search time, and file size according to the type of \texttt{FAISS} index. The index search time is the time in seconds that takes to retrieve top-k documents for a question using the index, averaged from the search on 100 questions. Two Xeon Gold 5120 CPU cores are used to measure the search time.}
    \label{table:faiss}
    \footnotesize
    \begin{threeparttable}
    \begin{tabular*}{1.0\textwidth}{l@{\extracolsep{\fill}}ccccccc}
        \toprule
  &\multicolumn{4}{c}{Retrieval Recall Rate} & \multicolumn{2}{c}{Index Search Time} & \\
  \cmidrule{2-5} \cmidrule{6-7}
                         & Top-1 & Top-20 & Top-50 & Top-100  & Top-1 & Top-10 & File Size \\
\midrule
IndexFlatIP                 & 54.2  & 82.8   & 86.3   & 88.2         & 29.8 & 33.9  & 61G\\
IndexHNSWFlat & -0.2  & -0.6   & -0.6   & -0.7    & 0.03 & 0.04     & 141G     \\
IndexHNSWSQ   & -0.2  & -0.7   & -0.6   & -0.7    & 0.23 & 0.24   & 96G    \\
        \bottomrule
    \end{tabular*}
    \end{threeparttable}
\end{table}

\paragraph{Batch Size and Number of Passages}
The original training scheme of DPR retriever uses in-batch negatives training with number of negatives of 127, which is a setup that utilizes all the passages for the other questions in the same batch as the negative passages for a question. \citet{karpukhin2020dense} show that such a training scheme is effective at boosting the performance of a two-tower retriever, that DPR retriever trained with and without in-batch negatives where the number of negatives is 7 shows a large gap of 8.5 in the retrieval recall rate at $k=5$.

On the other hand, the same setup cannot be applied for the training of reader; the reader is a one-tower model and thus needs to compute the score once for every the $(B \times B)$ pairs, unlike the two-tower retriever that can separately encode the questions and passages and use matrix multiplication to get the scores. Since the reader is the teacher to our retriever, RDR, it is trained with a smaller number of question-passage pairs at training time. As described in Section~\ref{subsec:training}, RDR is trained with batch size of 10 and 16 passages per question, but still shows a significant improvement in retrieval performance, especially at $k=1$.

\paragraph{FAISS Configuration}

To retrieve the top-k passages which becomes the input to the reader, we use the Maximum Inner Product Search (MIPS) index built with \texttt{FAISS}~\citep{8733051}.
For a fair comparison, the results of the experiments related to DPR are reported using IndexFlatIP as in the work of \citet{karpukhin2020dense}, and those related to RAG are reported using IndexHNSWFlat with 512 neighbors to store per node, construction time search depth of 200, search depth of 128 and L2 search metric with the max norm trick~\citep{bachrach2014speeding}. To prevent OOM while finetuning RAG reader, we additionally built and used IndexHNSWSQ index with 8bit scalar quantization. Table~\ref{table:faiss} shows the performance achieved with different types of \texttt{FAISS} index.

\subsection{The Effect of Reading Fewer Passages at Inference Time}
\label{appendix:latency}

\begin{table}[h]
    \setlength{\tabcolsep}{3.5pt}
    \centering
    \caption{Latency of DPR reader with respect to the number of passages to read. The inference time is averaged from the runs on 100 question-passage pairs of batch size 1. Two Xeon Gold 5120 CPU cores are used across the experiments. The latency linearly increases with respect to the number of passages.}
    \label{table:inference_time}
    \footnotesize
    \begin{threeparttable}
    \begin{tabular*}{0.8\textwidth}{l@{\extracolsep{\fill}}cccccc}
        \toprule
                        Device & Top-1 & Top-10 & Top-20 & Top-30 & Top-40 & Top-50 \\
\midrule
CPU & 0.63 & 6.76 & 13.49 & 20.17 & 28.9 & 36.1 \\
GPU (1 $\times$ P40) & 0.02 & 0.12 & 0.23 & 0.35 & 0.45 & 0.57 \\
        \bottomrule
    \end{tabular*}
    \end{threeparttable}
\end{table}

\subsection{Open-Domain QA Models}\label{appendix:sota}

\begin{table}[ht]
    \setlength{\tabcolsep}{3.5pt}
    \centering
    \caption{End-to-end QA (Exact Match) accuracy of recent and state-of-the-art models. Each of the results in the left and right columns of TriviaQA is the score on the open domain test set and hidden test set, respectively.}
    \label{table:sota}
    \footnotesize
    \begin{threeparttable}
    \begin{tabular*}{0.8\textwidth}{l@{\extracolsep{\fill}}ccc}
        \toprule
\textbf{Model}                     & \textbf{NQ-test} & \multicolumn{2}{c}{\textbf{TriviaQA-test}}      \\
\midrule
\midrule
Path Retriever~\citep{asai2019learning}            & 31.7    & -             & -    \\[0.2em]
Graph Retriever~\citep{min2019knowledge}          & 34.7    & 55.8          & -    \\[0.2em]
Hard EM~\citep{min2019discrete}                  & 28.8    & 50.9          & -    \\[0.2em]
ORQA~\citep{lee2019latent}                      & 31.3    & 45.1          & -    \\[0.2em]
REALM~\citep{guu2020realm}                     & 38.2    & -             & -    \\[0.2em]
DPR-Single~\citep{karpukhin2020dense}                & 41.5    & 56.8          & -    \\
$\drsh$ w/ RDR & 42.1 (+0.6)  & 57.0 (+0.2) \\[0.2em]
BM25+DPR-Multi~\citep{karpukhin2020dense}            & 38.8    & 57.9          & -    \\[0.2em]
SpanSeqGen~\citep{min2020ambigqa}                & 42.5    & -             & -    \\[0.2em]
RAG-Token~\citep{lewis2020retrieval}                 & 44.1    & 55.2          & 66.1 \\
$\drsh$ w/ RDR & 44.5 (+0.4) & -           & -  \\[0.2em]
RAG-Sequence~\citep{lewis2020retrieval}              & 44.5    & 56.1          & 68.0 \\[0.2em]
T5~\citep{roberts2020much}                        & 36.6    & -             & 60.5 \\[0.2em]
GPT-3 few shot~\citep{brown2020language}            & 29.9    & -             & 71.2 \\[0.2em]
Fusion-in-Decoder (base)~\citep{izacard2020leveraging}  & 48.2    & 65.0          & 77.1 \\[0.2em]
Fusion-in-Decoder (large)~\citep{izacard2020leveraging} & 51.4    & 67.6          & 80.1 \\
        \bottomrule
    \end{tabular*}
    \end{threeparttable}
\end{table}
In Table~\ref{table:sota}, we show the end-to-end QA accuracy of recent and state-of-the-art models along with that of the readers improved with RDR. Although we could not apply RDR to the current state-of-the-art, Fusion-in-Decoder~\citep{izacard2020leveraging}, due to the lack of computation resources\footnote{The performance of Fusion-in-Decoder (large) is obatined using 64 $\times$ 32G GPUs for training.} and public checkpoints, the observed improvements in EM with RDR for DPR-Single and RAG-Token suggest that a higher accuracy may be achieved with RDR.

\subsection{Ablation Study}\label{appendix:ablations}

\begin{table}[t!]
    \setlength{\tabcolsep}{3.5pt}
    \centering
    \caption{Ablation result which shows the drop in the end-to-end QA accuracy (Exact Match) when no finetuning is applied to the reader while the retriever is swapped with RDR.}
    \label{table:ablation_em}
    \footnotesize
    \begin{threeparttable}
    \begin{tabular*}{1.0\textwidth}{c@{\extracolsep{\fill}}lccc}
\toprule
                       \textbf{Dataset}   & \multicolumn{1}{c}{\textbf{Model}}       & \multicolumn{3}{c}{\textbf{Ablation Result}}            \\
\cmidrule{3-5}
               &                       & Top-1       & \multicolumn{2}{c}{Reported}            \\
\cmidrule{3-3} \cmidrule{4-5}
                                    &                       & EM          & EM          & Top-k       \\
\midrule
\midrule
\multirow{4}{*}{\textbf{NQ-test}}       & DPR-Single w/ RDR & 37.3        & 42.1        & 10          \\
                       & \quad w/o reader finetuning             & 37.2 (-0.1) & 40.9 (-1.2) & 10          \\
\cmidrule{2-5}
                       & RAG-Token w/ RDR & 40.9        & 44.5        & 15          \\
                       & \quad w/o reader finetuning             & 37 (-3.9)   & 42.9 (-1.6)  & 15          \\
\midrule
\multirow{2}{*}{\textbf{TriviaQA-test}} & DPR-Single w/ RDR & 49.1        & 57          & 50          \\
                       & \quad w/o reader finetuning             & 49.2 (+0.1) & 56.1 (-0.9) & 50         \\
        \bottomrule
    \end{tabular*}
    \end{threeparttable}
\end{table}

Table~\ref{table:ablation_em} reports the results of ablation studies that the reader is not finetuned while the retriever is swapped with RDR. A drop in EM is observed in general without finetuning the reader, which may have come from the input distribution gap between the training and inference phases.

\end{document}